\documentclass{article}

\usepackage{PRIMEarxiv}

\usepackage[utf8]{inputenc} 
\usepackage[T1]{fontenc}    
\usepackage{hyperref}       
\usepackage{url}            
\usepackage{booktabs}       
\usepackage{amsfonts}       
\usepackage{nicefrac}       
\usepackage{microtype}      
\usepackage{lipsum}
\usepackage{fancyhdr}       
\usepackage{graphicx}       
\graphicspath{{media/}}     
\usepackage{algorithm}
\usepackage{algorithmic}
\usepackage{longtable}
\usepackage{booktabs}
\usepackage{lineno}
\usepackage{multirow}
\usepackage{amsmath,amssymb,bm,amsthm}
\title{Gradient Surgery for Physics-Informed Neural Networks}

\author{
  Thomas Borsani, Giuseppe Di Fatta \\
  Faculty of Engineering \\
  Free University of Bozen - Bolzano \\
  Bozen - Bolzano, Italy\\
  \texttt{\{tborsani, giuseppe.difatta\}@unibz.it} 
}

\begin{document}
\maketitle

\begin{abstract}
Physics-Informed Neural Networks (PINNs) are trained by optimising a composite objective that combines data fitting with physics-based constraints, typically resulting in a highly imbalanced multi-task optimisation problem. Under these conditions, existing optimisation strategies are affected by conflicting task gradients, leading to slow convergence and unstable training, particularly for stiff and high-frequency partial differential equations. We analyse gradient conflicts throughout training of PINNs with standard optimiser and investigate Multi-Task Deep Learning (MTDL) optimisation methods. 
In our analysis across four benchmark problems we observed that PINN optimisation exhibits three distinct phases in which angle- and magnitude-based gradient conflicts alternate, with only one present at a time.
Building on these observations, we propose PAM-GS, a physics-aware gradient surgery method that adaptively mitigates task interference during training according to the observed conflict types. 
Experiments on four representative PDE benchmarks demonstrate that PAM-GS combines competitive solution accuracy with consistently strong task-balanced performance, outperforming existing methods on most problems.
Code is available at \href{https://github.com/ThomBors/gs_pinn}{https://github.com/ThomBors/gs\_pinn}
\end{abstract}


Recent advances in deep learning have driven growing interest in neural networks models for solving partial differential equations (PDEs) \cite{beck2023_DCDS,han2018_PNAS}. 
PDEs arise across diverse scientific disciplines, including fluid dynamics, electromagnetics, and financial mathematics \cite{Chen2020_opt,geneva2020_jcp,Jin2021_jcp}. 
However, solving nonlinear and high-dimensional PDEs efficiently remains a fundamental challenge for traditional numerical methods due to stability and computational constraints. Physics-Informed Neural Networks (PINNs) address this challenge by formulating PDE solving as the optimisation of a neural network constrained by the governing physical laws \cite{pinns2019_jcp,Cuomo2022_JSC}. 
Specifically, PINNs employ an implicit neural representation, where a neural network parametrises the solution field as a continuous, differentiable function, enabling spatial and temporal derivatives to be computed efficiently via automatic differentiation. 
The governing PDE residuals are enforced at collocation points within the computational domain, while boundary and initial conditions are included as additional terms in the training objective.

Despite its simplicity, this formulation gives rise to a challenging multi-objective optimisation problem \cite{Krishnapriyan2021_NIPS,Cuomo2022_JSC,Mario2023_ryoalSoc,Wang2022_jcp,Wang2021_siam}. 
Competing objectives can result in gradient imbalances caused by numerical stiffness \cite{Wang2021_siam}, heterogeneous convergence rates between loss components \cite{Wang2022_jcp}, the soft enforcement of physical constraints \cite{Krishnapriyan2021_NIPS}, poor initialisation \cite{Wong2024_tai}, and inefficient collocation sampling strategies. 
These factors often lead to slow convergence, optimisation stagnation, and reduced solution accuracy. 
Consequently, adaptive weighting schemes have been proposed to balance the loss functions for PDE residuals, boundary conditions, and initial conditions~\cite{levi2023_jcp}, although no consensus exists on an optimal strategy.

In parallel, Multi-Task Deep Learning (MTDL) addresses analogous issues via gradient surgery methods, which mitigate inter-task interference by modifying conflicting gradients during joint optimisation~\cite{yu2024_arxiv}. 
Such conflicts arise from misaligned gradient directions (angle-based) and/or disproportionate magnitudes that dominate shared updates (magnitude-based)~\cite{Yu2020NeurIPS_pcgrad}. 
Similar phenomena are expected in PINNs~\cite{liu2025_iclr}, where PDE residual gradients can dominate boundary and initial condition gradients~\cite{Wang2021_siam}, and where the residual loss landscape is often highly non-convex due to underdetermined PDE constraints~\cite{Daw2023_icml}.
However, a systematic study of gradient conflicts and a comprehensive evaluation of gradient surgery methods in this setting remains limited.

This work studies the training dynamics of PINNs, with a focus on the emergence of gradient conflicts~\cite{Yu2020NeurIPS_pcgrad} across canonical PDE benchmarks. Specifically, we consider problems that capture a range of behaviours, including non-linear convection, dispersion, and incompressible fluid dynamics, making them suitable testbeds for analysing optimisation behaviour in PINNs.
We perform a systematic evaluation of modern gradient surgery methods to assess their impact on optimisation behaviour.
Our empirical analysis reveals a distinct temporal structure in gradient conflicts: task-angle-based gradient conflict dominates early training, while later stages are primarily affected by magnitude gradient conflicts. 
Based on these findings, we propose Physics-Aware Momentum Gradient Surgery (PAM-GS), a conflict-aware gradient method for PINNs that outperforms existing optimisers in most cases.
Our main contributions are summarised as follows:
\begin{itemize}
    \item We study the emergence of gradient conflicts in PINNs across canonical 1D and 2D PDE benchmarks.
    \item We perform a systematic evaluation of modern gradient surgery methods and quantify their impact on optimisation behaviour.
    \item We propose Physics-Aware Momentum Gradient Surgery (PAM-GS), a conflict-aware optimisation method that improves performance over existing approaches.
\end{itemize}

\section{Physics-Informed Neural Networks}

\paragraph{Preliminaries.}
We consider the initial–boundary value problem for a general partial differential equation (PDE) defined over a scalar field $u(x,t): \Omega \times (0,T] \rightarrow \mathbb{R}$:
\begin{align}
\mathcal{N}[u(x,t), x,t] &:= \mathcal{N}[u(x,t), x,t] + f(x,t) = 0, \quad x \in \Omega,\ t \in (0,T], \\
\mathcal{B}[u(x,t), x,t] &:= \mathcal{B}[u(x,t), x,t] + g(x,t) = 0, \quad x \in \partial\Omega,\ t \in (0,T], \\
\mathcal{I}[u(x,0), x,0] &:= u(x,0) + h(x) = 0, \quad x \in \Omega,
\end{align}
where $\Omega \subset \mathbb{R}^d$ denotes the spatial domain with boundary $\partial\Omega$, and $\mathcal{N}$ and $\mathcal{B}$ are differential operators that encode the dynamics of PDE and the boundary conditions, respectively. The functions $f$, $g$, and $h$ represent the source terms.

\paragraph{PINN formulation.}
To solve this system, PINNs introduce a parametric neural approximation $\hat{u}(x,t;\theta)$ of the solution $u(x,t)$, where $\theta$ denotes the network parameters. 
The spatial and temporal derivatives required by the operators $\mathcal{N}$ and $\mathcal{B}$ are calculated using automatic differentiation applied to $\hat{u}(x,t;\theta)$.
The PDE is then enforced by minimising the following composite loss:
{\small
\begin{equation}
\label{formula:MTLloss}
\mathcal{L}(\theta)
=
\underbrace{\frac{1}{n_N}\sum_{i=1}^{n_N} \left\| \mathcal{N}[\hat{u}(x_i,t_i;\theta), x_i,t_i] \right\|}_{\mathcal{L}_{PDE}}
+
\underbrace{\frac{1}{n_B}\sum_{i=1}^{n_B} \left\| \mathcal{B}[\hat{u}(x_i,t_i;\theta), x_i,t_i] \right\|}_{\mathcal{L}_{BC}}
+
\underbrace{\frac{1}{n_I}\sum_{i=1}^{n_I} \left\| \mathcal{I}[\hat{u}(x_i,0;\theta), x_i,0] \right\|}_{\mathcal{L}_{IC}}.
\end{equation}
}

\paragraph{Optimisation structure.}
PINNs can be formulated as a MTDL problem, where a neural network with shared parameters ( $\theta \in \mathbb{R}^m $) jointly optimises ($K \geq 2$) physics-based objectives. 
Each physical constraint defines a task-specific loss, including the PDE residual ( $\mathcal{L}_{\mathrm{PDE}} $), boundary conditions ( $\mathcal{L}_{\mathrm{BC}} $), and initial conditions ( $\mathcal{L}_{\mathrm{IC}} $).
The resulting optimisation problem is given by $\theta^\ast = \arg\min_{\theta \in \mathbb{R}^m}\mathcal{L}(\theta).$

The  $\mathcal{L}_{\mathrm{PDE}} $ enforces the governing equations at collocation points throughout the spatio-temporal domain, whereas the $\mathcal{L}_{\mathrm{BC}} $ and $\mathcal{L}_{\mathrm{IC}} $ impose soft constraints on the solution at the spatial boundaries and the initial time, respectively. 
This decomposition explicitly exposes the multi-objective nature of PINN optimisation, where gradients interaction induced by different physical constraints may interfere, leading to conflicting optimisation directions.

\section{Conflicting Gradients in Multi-Task Deep Learning}
In MTDL, model parameters are optimised concurrently with respect to multiple task-specific objectives. 
Consequently, gradients associated with different tasks may interfere during optimisation, reducing convergence efficiency and degrading overall performance. 
Gradient conflicts can generally be categorised into two complementary forms.

\paragraph{Angle-Based Gradient Conflict.}
Consider two task-specific gradients \( g_i, g_j \in \mathbb{R}^d \). 
An angle-based conflict arises when the two update directions are misaligned such that their inner product becomes negative. 
Equivalently, this corresponds to an angle larger than \(90^\circ\), as expressed in Equation \eqref{eq:cosine similarity}.

\begin{equation}
\label{eq:cosine similarity}
    \cos(\phi_{ij}) = \frac{g_i \cdot g_j}{\|g_i\| \|g_j\|} < 0.
\end{equation}

From an optimisation perspective, such misalignment implies that the resulting aggregated update cannot simultaneously reduce both objectives along a shared descent direction. 
When gradients are weakly misaligned (e.g., close to orthogonal), progress is still partially inhibited, since the effective descent component is reduced. 
In the extreme case of opposing gradients, updates cancel, leading to vanishing progress for both tasks.

\paragraph{Magnitude-Based Gradient Conflict.}
Let \( g_i, g_j \in \mathbb{R}^d \) be task gradients. 
A magnitude-based conflict refers to situations where one task gradient dominates the update due to a substantially larger norm. 
To quantify this imbalance, we use:

\begin{equation}
\label{eq:magnitude similarity}
\psi(g_i, g_j) = \frac{2 \|g_i\|_2 \|g_j\|_2}{\|g_i\|_2^2 + \|g_j\|_2^2}.
\end{equation}

Low values of \( \psi(g_i, g_j) \) indicate a strong imbalance between tasks, which can bias parameter updates toward goals with larger gradient magnitudes. 
Unlike Angle-Based Gradient Conflicts, magnitude imbalance does not necessarily imply opposing objectives; instead, it reflects disparities in scaling induced by loss geometry, normalisation choices, or differing optimisation difficulty across tasks.

\paragraph{Gradient Surgery Methods.}
To mitigate gradient interference in MTDL, Gradient Surgery methods define heuristic procedures that modify task-specific gradients before aggregation. 
Formally, for \(K\) tasks, the layer \(l\) update is computed via an aggregation operator:
\(
\nabla_{\theta^{(l)}} \mathcal{L}
= s\big(
\nabla_{\theta^{(l)}} \mathcal{L}_1,\dots,
\nabla_{\theta^{(l)}} \mathcal{L}_K
\big),
\)
where \( s(\cdot) \) defines the specific combination rule induced by a given gradient surgery strategy.

\section{Related Work}
Existing approaches to mitigating gradient conflicts fall into two main categories.

\paragraph{Loss Balancing.}
Loss balancing methods aim to appropriately weight the loss functions of the tasks involved in a joint optimisation setting. 
A range of strategies have been introduced to estimate suitable weights for each task. 
The \textbf{UW} method \cite{Kendall2018_uw} employs the homoscedastic uncertainty associated with each task to determine the corresponding loss weights. 
In contrast, \textbf{UW-SO} \cite{Kirchdorfer2025_pr} uses the inverse of the task-specific losses, which demonstrate to be the analytically optimal UW scheme.
Meanwhile, \textbf{DWA} \cite{Liu2019_dwa} derives the task weights from the temporal rate of change of the individual task losses.
\textbf{GradNorm} \cite{Chen2018_gradnorm} adjusts the weights according to the magnitude of the corresponding gradients. 
Unlike these schemes, \textbf{RLW} \cite{Lin2022_rgw} simply samples random weights. 
Furthermore, \textbf{FAMO} \cite{Liu2023NeurIPS_famo} learns the weights dynamically based on the effectiveness of the loss updates. 
Collectively, these techniques reduce gradient interference by preventing any task from dominating training.

\paragraph{Gradient Surgery.}
These methods accelerate MTDL convergence by reweighting task gradient components. 
They introduce heuristics that reduce conflicts and guide optimisation. 
\textbf{Nash-MTL} \cite{Navon2022_nashmtl} uses the Nash Bargaining Solution to balance task gradients, while \textbf{MGDA} \cite{DESIDERI2012_mgda,Dong2015_mgda_translation} finds a direction that jointly decreases all objectives under multi-objective KKT \cite{KuhnTucker1951_kkt} conditions. 
Although computationally demanding, these approaches are effective.

Other methods more directly target gradient conflicts. 
\textbf{GDOD} \cite{Dong2023_godo} decomposes task gradients into shared and conflicting parts, updating only the shared ones.
\textbf{PCGrad} \cite{Yu2020NeurIPS_pcgrad} decorrelates gradients to reduce conflicts, and \textbf{CAGrad} \cite{Liu2021_cagrad} finds a conflict-averse update direction. 
\textbf{GradDrop} \cite{Chen2020_graddrop} enforces consistent gradient signs across tasks. 
\textbf{IMTL} \cite{Liu2021_imtl} finds a path along which cosine similarities between task gradients stay stable, and \textbf{Aligned-MTL} \cite{Senushkin2023_alignedmtl} aligns the principal components of the gradient matrix. 
These methods often outperform the more complex Nash-MTL \cite{Navon2022_nashmtl} while keeping computational costs low. \textbf{FairGrad} \cite{Hao2024_fairgrad} frames gradient aggregation as an $\alpha$-fair utility optimisation problem to balance task improvements and reach Pareto-stationary solutions. 
\textbf{CONFIG} \cite{liu2025_iclr} is tailored for PINNs and handles gradients by building a normalised task-gradient space and computing updates via a pseudoinverse-based projection. 
This enforces consistent alignment of loss components so all physics constraints contribute positively to the shared update, preventing any single term from dominating.
\textbf{SAM-GS} \cite{borsani2026_ecml} identifies magnitude-based gradient conflicts as a key source of sub-optimal optimisation in MTDL settings, and proposes a momentum-based optimisation to mitigate their effect by stabilising gradient magnitudes across tasks.

\section{Comparative Analysis Methodology}
In this section, we present the metrics and benchmarks used to conduct the empirical analysis on gradient conflicts and the optimisation behaviour of surgery methods for PINNs.

\paragraph{Benchmarks.}
We evaluate the proposed methods on four canonical PDE benchmarks spanning increasing dimensionality and diverse physical regimes: the 1D unsteady Burgers’ equation, the 1D unsteady Schrödinger equation, the 2D Kovasznay flow (Navier--Stokes equations), and the 3D unsteady Beltrami flow (Navier--Stokes equations). 
Further details are provided in Appendix~\ref{appendix:PDE specification}.
These problems capture non-linear convection, dispersive dynamics, and incompressible fluid behaviour, providing a diverse testbed for analysing optimisation dynamics in PINNs.

\paragraph{Gradient surgery methods.}
To study the effect of gradient interference on optimisation dynamics, we developed an experimental protocol to run, test and compare many representative MTDL optimisation methods.
The selected methods include classical reweighing strategies such as DWA, UW-SO and FAMO, as well as a wide range of gradient manipulation techniques.
Within gradient surgery methods, we consider PCGrad, GradDrop, IMTL, CAGrad, Aligned-MTL, CONFIG, Nash-MTL, and SAM-GS. 

\paragraph{Performance Metrics}
We evaluate gradient surgery methods on the test set using the Mean Squared Error (MSE) at the initial condition (IC), boundary condition (BC), and PDE residual (PDE) for each governing equation, together with the overall test MSE. 
To provide an aggregate comparison across metrics, we report the Mean Rank (MR) and the relative improvement ($\bm{\Delta M\%}$)~\cite{Navon2022_nashmtl}. 
MR is computed by averaging the rank of each method across all evaluation metrics. 
$\Delta M\%$ measures the relative improvement over a baseline method. 
Lower values of MR and $\Delta M\%$ indicate better performance.

\begin{align}
    \Delta M \% = \frac{1}{K} \sum_{k=1}^K \frac{m_{\mathrm{MTL},k} - m_{\mathrm{base},k}}{m_{\mathrm{base},k}} \cdot 100
    \label{formula:deltam}
\end{align}

where $m_{\mathrm{MTL},k}$ and $m_{\mathrm{base},k}$ denote the performance metrics of the MTDL optimisation method and the vanilla joint training method for task $k$, respectively.

\section{Addressing Gradient Conflicts in PINNs}
This section empirically studies gradient conflicts during PINN training and evaluates how different gradient surgery methods affect their structure and frequency.
To quantify gradient conflicts, we compute two measures during training of vanilla PINNs using the Adam optimiser.
The average cosine gradient angle \eqref{eq:cosine similarity}, which captures task gradient alignment and serves as a proxy for interference, and the average gradient magnitude ratio \eqref{eq:magnitude similarity}, which measures scale imbalance across tasks.

\begin{figure}[tpb]
    \centering
    \includegraphics[width=.8\linewidth]{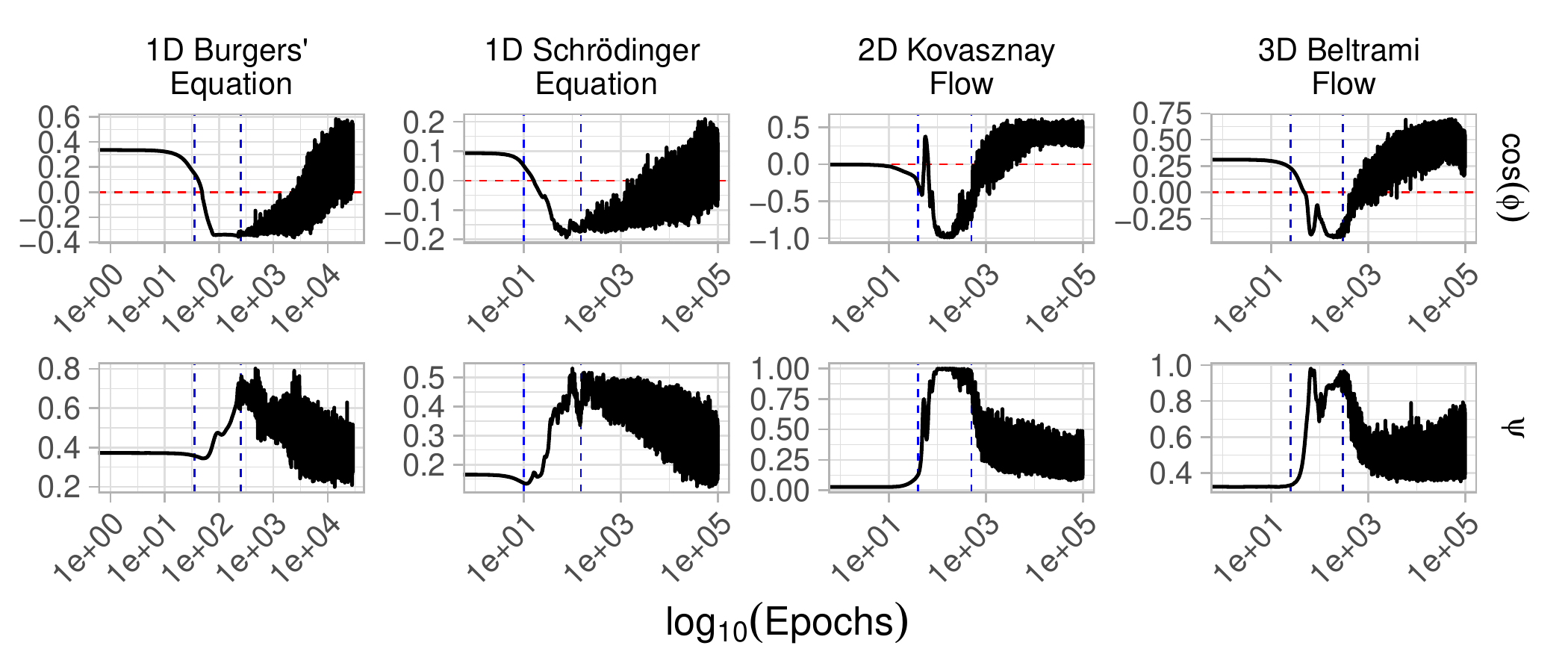}
    \caption{Evolution of gradient conflict metrics ($cos(\phi)$,$\Psi$) during PINNs training with ADAM optimiser. The red line denotes the angle-based gradient conflict threshold, while the blue lines highlight the transitions between training phases, from Fitting phase to Diffusion and to Total Diffusion.}
    \label{fig:Conflicting_Transition}
\end{figure}

Figure~\ref{fig:Conflicting_Transition} shows that both angle- and magnitude-based gradient conflicts align closely with the Fitting–Diffusion–Total Diffusion learning phases identified in~\cite{Anagnostopoulos2026_nn}, despite those phases originally being defined in terms of inter-batch gradient homogeneity. Here, the same phase structure is observed in the interaction of gradients arising from different loss terms within a single batch, transitioning from heterogeneous to progressively aligned and equilibrium-like regimes in total diffusion. 
These regimes correspond to distinct transitions in optimisation dynamics and are consistent with the evolution of gradient conflicts throughout training.
During the Fitting phase, angle‑based conflicts are negligible, whereas gradient magnitudes exhibit low similarity across batches. 
The subsequent Diffusion phase is characterised by increasing angular conflict alongside a rise in magnitude similarity. 
In the Total Diffusion phase, angular conflict decreases (indicating stronger directional alignment), but magnitude consistency deteriorates, while remaining higher than in the initial phase. 
These findings suggest the emergent phase structure is not solely due to batch variability.

Motivated by this, we evaluate existing gradient surgery methods. 
Most approaches couple angular and magnitude conflicts, ignoring the three-phase learning structure and thereby potentially degrading optimisation performance \cite{borsani2026_ecml}.
SAM-GS is among the few methods that explicitly detects gradient conflicts \cite{borsani2026_ecml}, but it implicitly assumes co-occurrence of angular and magnitude conflicts. 
In regimes where magnitude imbalance dominates without strong angular disagreement, similarity-based updates may still be applied, which can be harmful in the ill-conditioned loss landscapes typical of PINNs \cite{Krishnapriyan2021_NIPS}.

\paragraph{Physics-Aware Momentum Gradient Surgery.}
Building on these observations, we introduce Physics-Aware Momentum Gradient Surgery (PAM-GS), a new gradient descent optimiser that explicitly detects angular and magnitude conflicts and adjust the gradient accordingly. When angle-conflicts are detected the approach introduce a modulation of the overall gradient with a momentum based on the gradient-magnitude similarity while for magnitude-based conflicts task gradient equalisation is adopted. 
In Algorithm~\ref{algorithm:pam-gs}, let $\theta_t$ denote the model parameters and $\mathbf{g}_{k,t}=\nabla_{\theta}\mathcal{L}_k(\theta_t)$ the gradient of task $k$, $k=1,\ldots,K$. 
At each iteration, PAM-GS computes the magnitude similarity $\Psi_t$~\eqref{eq:magnitude similarity} and the cosine similarity $\Phi_t$~\eqref{eq:cosine similarity} by aggregating pairwise similarities among the $d$ task gradients per hidden layer, where $d$ is the number of hidden units. 
By averaging over the $K$ tasks the type of conflict can be detected.
In particular, when $\Psi_t<\gamma$ a magnitude conflict is detected as gradient magnitudes are highly imbalanced. Under this condition, task gradients with larger norm would dominate the update, inducing a bias towards their objectives. 
Therefore, in order to mitigate this problem, PAM-GS equalises task gradients before their aggregation. 
When $cos\left(\Phi_t\right)<0$, an angle-based conflict is detected, 
the task gradients partially cancel and the magnitude of their sum decreases. This would slow the training process, though it may not affect the overall gradient direction. 
PAM-GS addresses this regime through momentum-based reweighing, with $\mathbf{m}_{k,t}$ tracking task-gradient dynamics and $h_t$ tracking $(1-\Psi_t)^2$. The coefficients $\beta_1$ and $\beta_2$ control the corresponding exponential moving averages, and $\hat{\cdot}$ denotes their bias-corrected estimates. 
The resulting weights exploit temporal consistency, with high magnitude similarity across recent updates permitting stronger momentum, whereas persistent dissimilarity increases regularisation and enforces more conservative updates.
If neither conflict is detected, PAM-GS directly aggregates the task gradients, where $\mathbf{1}_d$ denotes the $d$-dimensional all-ones vector.

\begin{algorithm}[t]
   \caption{Physics-Aware Momentum Gradient Surgery}
   \label{algorithm:pam-gs}
\begin{algorithmic}
    \STATE {\bfseries Hyperparameters:} $\beta_1 \leftarrow 0.9, \beta_2 \leftarrow 0.99, \gamma \leftarrow 0.1$ 
    \STATE {\bfseries Initialise:} $\theta_0, m_0 \leftarrow 0 , h_0 \leftarrow 0, t \leftarrow 0, \epsilon \leftarrow 1e-8 $
    \REPEAT
    \STATE $t \leftarrow t + 1$
    \STATE $\mathbf{g}_k \leftarrow \nabla_{\theta_t} \mathcal{L}_k, \forall k $
    \STATE $\Psi_t = \frac{1}{K} \sum_{k=1}^{K} \left( \frac{1}{d^2} \sum_{i=1}^{d}\sum_{j=1}^{d} \psi(g_{k,i,t},g_{k,j,t}) \right); \quad \Phi_t = \frac{1}{K} \sum_{k=1}^{K} \left( \frac{1}{d^2} \sum_{i=1}^{d}\sum_{j=1}^{d} \phi_{ij,t} \right)$
    \STATE $\vec{m}_{k,t} \leftarrow \beta_1 \vec{m}_{k,t-1} + (1 - \beta_1) \mathbf{g}_k, \forall k ; \quad h_t \leftarrow \beta_2 h_{t-1} + (1 - \beta_2)(1 -\Psi_t)^2 $
    \STATE $\widehat{\vec{m}}_{k,t} \leftarrow \frac{\vec{m}_{k,t}}{1-\beta_1 ^t}; \quad \widehat{h}_t \leftarrow \frac{h_t}{1-\beta_2 ^t}$
    \IF{$\Psi_t < \gamma$}
    \STATE $\mathbf{w}_k = \frac{ \overline{ \|\mathbf{g}_{k}\|}_2}{\|\mathbf{g}_{k}\|_2} \mathbf{1}_k, \forall k$ 
    \ELSIF{$cos\left(\Phi_t\right) < 0$}
    \STATE $ \mathbf{w}_k = \frac{|\widehat{\vec{m}}_{k,t}|} {\sqrt{\widehat{h}_t}+\epsilon}, \forall k$
    \ELSE 
    \STATE $ \mathbf{w}_k = \mathbf{1}_k$
    \ENDIF
    \STATE {\bfseries Update:} $\theta_t = \theta_{t-1} - \alpha \sum_{k=1}^{K} \mathbf{w}_k \odot \mathbf{g}_k$
   \UNTIL{convergence}
\end{algorithmic}
\end{algorithm}

The proposed algorithm therefore applies different update strategies according to the conflict type.
It prioritises the detection of magnitude-based gradient conflicts, as these have been reported to substantially hinder training performance, particularly during the fitting phase \cite{Jastrzebski2020_iclr}. While angle-based gradient conflicts can decelerate the convergence process \cite{borsani2026_ecml}.

\section{Performance Evaluation}\label{sec:experimentresults}
This section evaluates gradient surgery methods in terms of predictive performance for PINNs. 
All methods, including the baselines, are independently trained and results averaged across five different random weight initialisations. Relevant parameters are selected via hyperparameter optimisation, as detailed in Appendix~\ref{appendix:Training Details}.
\begin{table}
\centering
\footnotesize
\tiny 
\setlength{\tabcolsep}{2.5pt} 
\renewcommand{\arraystretch}{0.85} 
\begin{tabular}{lllllllllll}
\toprule
 &  & \multicolumn{3}{c}{Constraint-wise MSE} &  & \multicolumn{2}{c}{Overall MSE} &  &  &  \\ \cline{3-5} \cline{7-8}
GS method &  & IC & BC & PDE  &  & Mean & SD &  & MR & $\Delta M \%$ \\ \midrule
Base Line &  & 1.289 & 0.716 & 1.615 &  & 1.599 & 0.035 &  & 0 & 0 \\ \midrule
FAMO &  & 4993.946 & 1270.649 & 3719.593 &  & 3698.4728 &1.087&  & 11.0 & $>1000$ \\
UW-SO &  & 5071.515 & 1348.557 & 3797.431 &  & 3776.3081 & 66.17 &  & 12.0 & $>1000$ \\
IMTL &  & 18.171 & 850.346 & 583.555 &  & 581.59 &681.7&  & 10.0 & $>1000$ \\
Aligned-MTL &  & 1.314 & 0.758 & 1.706 &  & 1.689 &0.081& & 8.5 & 4.73 \\
SAM-GS & & 1.293 & 0.731 & 1.692 & &   1.675 &0.086&  &  8.0 &  2.94\\
DWA &  & 1.323 &  0.729 &  1.638 &  &  1.622 &0.071&  &  7.25 &  1.796 \\
GraDrop &  & 1.296 & 0.72 & 1.628 &  & 1.612 &0.045&  & 6.0 & 0.65 \\
CONFIG &  &  1.261 & 0.729 & 1.626  &  & 1.610 & 0.051 &  & 5.0 & 0.2 \\
Nash-MTL &  & 1.289 & 0.716 & 1.615 &  & 1.6 & 0.022&  & 4.0 & 0 \\
PCGrad &  & 1.229 & 0.717 & 1.607 &  & 1.598 & 0.063&  & 2.5 & -1.42 \\
CAGrad &  & \textbf{1.227} & \textbf{0.709} & \textbf{1.578} &   & \textbf{1.563} & 0.034& & \textbf{1.0} & \textbf{-2.62} \\
\midrule
PAM-GS &  & 1.237 & 0.715 & 1.599 &  & 1.583 &0.042&  & 2.75 & -1.82 \\
\bottomrule
\end{tabular}
\caption{Performance comparison of gradient surgery methods on the 1D Burgers equation. All MSE values are reported in units of $10^{-4}$.}
\label{table:1dburger}
\end{table}

\begin{table}
\centering
\resizebox{\textwidth}{!}{ 
\begin{tabular}{llcccccccccccccccccccc}
\toprule
&&
\multicolumn{4}{c}{Solution $u$} &&
\multicolumn{4}{c}{Solution $v$} &&
\multicolumn{4}{c}{Magnitude $h$} &&
\multicolumn{2}{c}{Overall MSE} && & \\
\cmidrule(lr){3-6}
\cmidrule(lr){8-11}
\cmidrule(lr){13-16}
\cmidrule(lr){18-19}
GS method &&
BC & IC & PDE & MSE &&
BC & IC & PDE & MSE &&
BC & IC & PDE & MSE &&
Mean & SD &&
MR & $\Delta$M (\%) \\
\midrule
Base Line &  & 1.269 & 1.770 & 35.86 & 35.42 &  & 0.306 & 0.566 & 55.62 & 54.92 &  & 1.049 & 1.203 & 3.662 & 3.630 &  & 31.32&4.288 &  &  0 & 0 \\
\midrule
UW-SO &  & 39430 & 854.5 & 44340 & 44130 &  & 1042 & 1568 & 33490 & 33080 &  & 37390 & 1522 & 67410 & 66890 &  & 48030 &114.3&  & 111.6 & $>1000$ \\
FAMO &  & 33950 & 3409 & 38650 & 38470 &  & 15.70 & 35.13 & 30440 & 30050 &  & 33800 & 3049 & 59890 & 59440 &  & 42650 &85.74&  & 10.73 & $>1000$ \\
IMTLg &  & 433.9 & 377.8 & 47040 & 46450 &  & 566.6 & 299.3 & 36590 & 36130 &  & 40.16 & 416.8 & 6005 & 5934 &  & 29500 & 381.3&  & 10.67 & $>1000$ \\
Aligned-MTL &  & 1.515 & 2.699 & 9350 & 9512 &  & 1.081 & 1.748 & 13710 & 13540 &  & 0.874 & 1.208 & 679.0 & 670.3 &  & 7907 & 112.8 &  & 8.8 & $>1000$ \\
DWA &  & 14.22 & 2.538  & 2045  &  2019 &  &  1.251 & 2.186 & 3049 & 3010 &  & 13.89 & 1.805 & 205.0 & 202.5 &  &  1744  &538.4&  & 8.200 &  $>1000$ \\
Nash-MTL &  & 1.335 & 1.719 & 36.60 & 36.15 &  & 0.320 & 0.603 & 56.49 & 55.78 &  & 1.116 & 1.171 & 3.788 & 3.754 &  & 31.89 &4.386&  & 6.73 & 2.471 \\
SAM-GS & & 0.647 & 1.468& 18.54  & 18.31 & & 19.20  & 0.425  & 38.18  & 37.70  & & 0.595 &  1.485  &2.237  & 2.219 & &  19.41 &4.960 & &  5.00  &-33.28\\
GradDrop &  & 0.89 & 1.096 & 20.78 & 20.53 &  & 0.267 & 0.469 & 31.62 & 31.22 &  & 0.738 & 0.813 & 2.265 & 2.245 &  & 18.00 &1.041&  & 5.07 & -35.60 \\
CONFIG &  &  0.810 & 1.309 & 9.216 &9.111 & &  0.181 & 0.411 & 17.38 & 17.16 & & 0.75 &   1.288 & 1.053 &  1.051 & & 9.107& 2.092& & 4.13 & -52.85 \\
PCGrad &  & 0.699 & 1.066 & 4.774 & 4.724 &  & 0.223 & 0.401 & 7.416 & 7.325 &  & 0.598 & 0.766 & \textbf{0.672} & \textbf{0.672} &  & 4.240 &0.427&  & \textbf{2.27} & -65.95 \\
CAGrad &  & 1.423 & 1.799 & \textbf{4.471} & \textbf{4.434} &  & 0.134 & 0.285 & \textbf{6.804} & \textbf{6.719} &  & 1.121 & 1.224 & 0.741 & 0.747 &  & \textbf{3.967} &0.18&  & 2.53 & -57.00 \\
\midrule
PAM-GS &  & \textbf{0.473} & \textbf{1.009} & 5.772 & 5.706 &  & \textbf{0.041} & \textbf{0.102} & 11.1 & 11.00 &  & \textbf{0.214} & \textbf{0.398} & 0.693 & 0.687 &  & 5.799 & 1.360&  & \textbf{2.27} & \textbf{-77.01} \\
\bottomrule
\end{tabular}
}

\caption{Performance comparison of gradient surgery methods on the 1D Schrödinger equation. All MSE values are reported in units of $10^{-5}$.}
\label{table:1dSchrodinger}
\end{table}
\paragraph{1D Burgers' equation (3 tasks)}
The results on the 1D Burgers' equation (Table~\ref{table:1dburger}) highlight the limitations of loss balancing methods, particularly more recent dynamic variants. 
The results shows that PAM-GS improves over SAM-GS and is competitive with the best results achieved by CAgrad for this benchmark.
Although CAGrad achieves a lower MSE, the performance of CAGrad and PAM-GS is comparable with the difference falling within the observed variability (SD).

\begin{table}
\centering
\tiny 
\setlength{\tabcolsep}{2.5pt} 
\renewcommand{\arraystretch}{0.85}
\begin{tabular}{lllllllllllllll}
\hline
 & \multicolumn{1}{c}{} & Pressure $p$ & \multicolumn{1}{c}{} & Velocity $u$ & \multicolumn{1}{c}{} & Velocity $v$ & \multicolumn{1}{c}{} & Velocity $w$ &  &\multicolumn{2}{c}{Overall MSE}  &  &  &  \\ \cline{3-3} \cline{5-5} \cline{7-7} \cline{9-9} \cline{11-12}
GS method &  & MSE &  & MSE &  & MSE &  & MSE &  & Mean & SD &  & MR & $\Delta M$ (\%) \\ \hline
Base Line &  & 1.946e+01 &  & 3.211 &  & 3.571 &  & 2.897 &  & 7.284 &1.505&  & 0 & 0 \\ \hline
UW-SO &  & 602200 &  & 95830 &  & 107200 &  & 98950 &  & 226000 &8734&  & 10.90 & $>1000$ \\
FAMO &  & 513200 &  & 85500 &  & 87890 &  & 86990 &  & 193400 &25.08&  & 10.10 & $>1000$ \\
DWA &   & 32.48 &  & 6.318 &  & 7.212 &  &  6.321 &  & 13.08 & 2.953& & 9.75 & 92.50\\
IMTLg &  & 33.46 &  & 3.823 &  & 3.524 &  & 4.218 &  & 11.26 &0.135&  & 8.1 & 36.71 \\
Nash-MTL &  & 20.67 &  & 2.942 &  & 3.168 &  & 2.967 &  & 7.436 &0.136&  & 6.2 & -2.491 \\
GradDrop &  & 18.31 &  & 3.787 &  & 4.097 &  & 3.074 &  & 7.316 &0.245&  & 5.75 & 4.811 \\
PCGrad &  & 16.45 &  & 3.563 &  & 4.059 &  & 4.080 &  & 7.038 &0.116&  & 6.5 & 9. \\
CONFIG &  & 18.56 &  & 2.477 &  & 3.951 &  & 2.209 &  & 6.799&0.435 &  & 5.2 & -6.791 \\
CAGrad &  & 19.04 &  & 2.510 &  & 2.265 &  & 2.624 &  & 6.609 & 0.052&  & 4.0 & -16.89 \\
SAM-GS &  & 16.64 &  & 2.016 &  & 2.397 &  & 2.980 &  & 6.009 &0.156&  & 3.5 & -18.36 \\ 
Aligned-MTL &  & \textbf{11.76} &  & 2.896 &  & 3.360 &  & 3.409 &  & 5.355 &0.078&  & 3.15 & -11.79 \\
\hline
PAM-GS &  & 14.14 &  & \textbf{1.602} &  & \textbf{1.762} &  & \textbf{1.273} &  & \textbf{4.693} & 0.073 &  & \textbf{1.200} & \textbf{-46.13} \\ \hline
\end{tabular}
\caption{Performance comparison of gradient surgery methods on the 3D Beltrami flow. All MSE values are expressed in units of $10^{-5}$.}
\label{table:3dbeltrami}
\end{table}

\paragraph{1D Schrödinger equation (3 tasks)}
The 1D Schrödinger equation considers a spatio-temporal problem defined over $(x,t) \in \mathbb{R}^2$. The PINN model predicts three scalar fields: the real component $u(x,t)$, the imaginary component $v(x,t)$, and the solution magnitude $h(x,t)$.
Results are reported separately for each component (Table~\ref{table:1dSchrodinger}), reflecting both phase and amplitude accuracy of the learned solution.
Although PAM-GS ranks third in overall MSE performance, it achieves the best aggregate relative improvement ($\Delta M \% = -77.01$) compared with ($\Delta M \% = -57.00$) for the lowest overall MSE method, demonstrating better performance across the full set of metrics.

\paragraph{3D Beltrami flow (3 tasks)}
The 3D Beltrami flow considers a spatio-temporal PINN defined over a 3D spatial domain with three velocity components $(u,v,w)$ and a pressure field $p$. The model predicts four scalar fields.
Results in Table~\ref{table:3dbeltrami} are reported using the MSE of each field variable, together with its decomposition across the IC, BC, and PDE residual terms. 
\begin{table}
\centering
\tiny 
\setlength{\tabcolsep}{1.5pt} 
\renewcommand{\arraystretch}{0.65}
\begin{tabular}{llllllllll}
\toprule
 &  & \multicolumn{2}{c}{Constraint-wise MSE} &  & \multicolumn{2}{c}{Overall MSE} &  &  &  \\ \cline{3-5} \cline{7-8}
GS method  &  & BC & PDE  &  & Mean &SD &  & MR & $\Delta M \%$ \\ \midrule
Base Line         &  & 7.347 & 5.041 &  & 5.044 &3.911&  & 0        & 0  \\
\midrule
UW-SO      &  & 20530000 & 4754000 &  & 4770000 &1857&  & 11.0 & $>1000$  \\
FAMO       &  & 5686000 & 2407000 &  & 2410000 &1978&  & 10.0 & $>1000$  \\
CONFIG     &  & 20.41 & 20.23 &  & 20.23 &3.77 &  & 10.0 & 2.62\\
SAM-GS     &  & 1.452 & 3.764 & &  3.762 &0.345&  & 7.3 & -15.51 \\
Nash-MTL    &  & 8.193 & 2.501 &  & 2.507 & 0.401&  & 7.7 & -29.73 \\
CAGrad     &  & 1.153 & 2.460 &  & 2.459 &0.105&  & 6.3 & -62.25 \\
IMTLg      &  & 4.975 & 2.244 &  & 2.247 & 0.145&  & 6.3 & -47.74 \\
PCGrad     &  & 2.975 & 1.938 &  & 1.939 &0.114 &  & 5.3 & -60.88 \\
DWA        &  & 2.142 & 1.595 & & 1.595 &0.069&  & 4.3 & -57.80 \\
GradDrop   &  & 8.275 & 1.333 &  & 1.340 & 0.127&  & 4.7 & -44.78 \\
Aligned-MTL &  & 0.719 & 0.521 &  & 0.522 &0.078 &  & 2.0 & -89.84 \\
\midrule
PAM-GS &  & \textbf{0.199} & \textbf{0.323} &  & \textbf{0.323} & 0.062&  & \textbf{1.0} & \textbf{-94.83} \\
\bottomrule
\end{tabular}
\caption{Performance comparison of gradient surgery methods on the 2D Kovasznay flow. All MSE values are reported in units of $10^{-7}$. }
\label{table:2dkovasznay}
\end{table}

For the sake of completeness, the full results are provided in Appendix~\ref{appendix:3DBeltrami}.
PAM-GS consistently achieves the strongest performance across most metrics and constraint categories in this setting, as reflected by its \textbf{MR} values approaching 1.

\begin{figure}[t]
    \centering
    \includegraphics[width=0.8\linewidth]{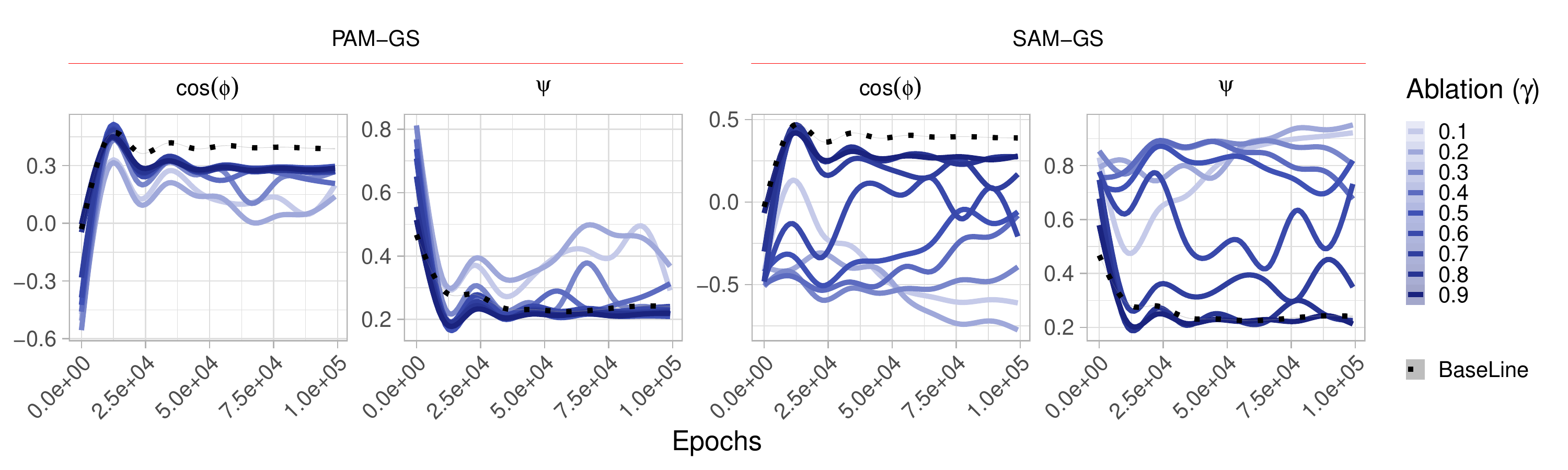}
    \caption{Ablation study of the parameter $\gamma$ for SAM-GS and PAM-GS gradient surgery methods during training on the 2D Kovasznay flow. Curves correspond to generalised additive model fits; shaded regions indicate 95\% confidence intervals.}
    \label{fig:kovasznayAblation}
\end{figure}

\paragraph{2D Kovasznay flow (2 tasks)}
The results on the 2D Kovasznay flow (Table~\ref{table:2dkovasznay}) are reported in terms of MSE, including contributions from boundary conditions and the PDE residual. 
PAM-GS consistently achieves the best overall performance across constraints, demonstrating improved accuracy in both components, as reflected by its \textbf{MR} of 1.
For this PDE benchmark, in Figure~\ref{fig:kovasznayAblation} we provide an analysis of the gradient conflicts with a comparison between SAM-GS and PAM-GS at varying parameter $\gamma$.
In SAM-GS low $\gamma$ values substantially degrade training dynamics due to excessive gradient modification under the implicit assumption of angular conflict.
In PAM-GS, the surgery interventions are properly adapted to the PINNs training phases, leading to a consistent and stable improvement compared to SAM-GS.

\begin{figure}
\centering
\includegraphics[width=0.8\linewidth]{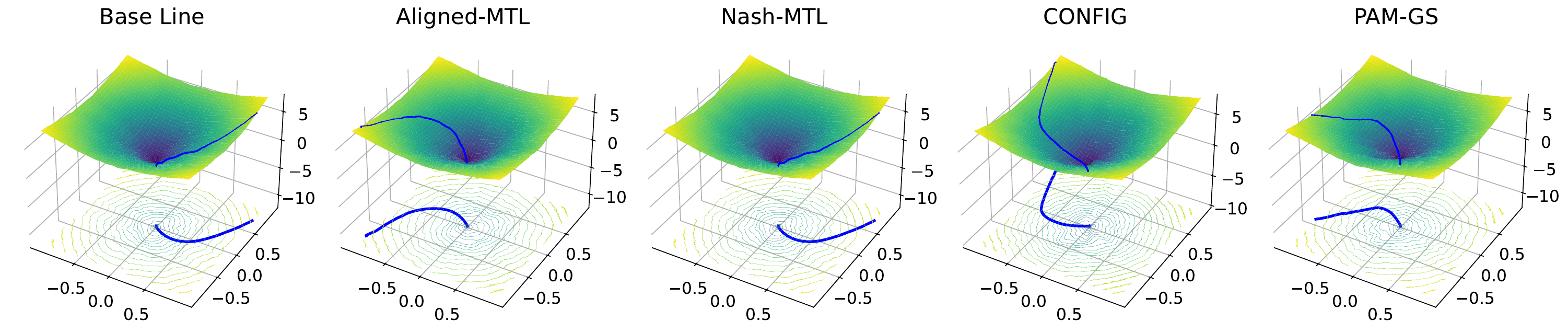}
\caption{Loss landscape of the 2D Kovasznay flow across optimisers. Panels show $\log_{10}$ training loss over a 2D PCA subspace with iso-contours and learning trajectories (blue) on the loss 3D surface and 2D projection. }
\label{fig:3DlosskovasznayMTLTRJ}
\end{figure}
We analyse the optimisation dynamics on the 2D Kovasznay flow via loss landscape visualisation. 
Following \cite{Li2018_nips}, we project optimisation trajectories and the final trained solution onto a 2D PCA subspace of the parameter space, and construct a loss surface by evaluating the loss over a regular grid in this projected space. 
The resulting contour plot is obtained by interpolating iso-loss level sets over the grid, providing a qualitative view of how optimisers traverse the underlying objective geometry.

Figure~\ref{fig:3DlosskovasznayMTLTRJ} shows that different optimisers exhibit distinct trajectories. 
The differences reflect varying behaviour in flat regions, sharp valleys, and high-curvature areas of the loss landscape. 
PAM-GS yields the shortest trajectory.

\section{Discussion}
Considering the overall MSE, PAM-GS performs particularly well on Kovasznay and Beltrami flows and remains among the top three methods on Burgers and Schrödinger. 
Among the strongest baselines, CAGrad performs best on Burgers and Schrödinger but ranks lower on Kovasznay (top-7) and Beltrami (top-4), whereas Aligned-MTL exhibits the opposite trend. 
In contrast, PAM-GS remains consistently competitive across all four benchmarks, suggesting that its PINN-specific, conflict-aware design is less sensitive to problem-dependent conflict patterns.
Given the multi-objective nature of PINNs, MSE alone does not capture task balance. 
\textbf{MR} and $\Delta M\%$ complement MSE by taking into account the relative performance across tasks. 
Under these metrics, PAM-GS achieves the best results on three of four benchmarks and remains competitive (top-2) on Burgers. 
It also consistently outperforms SAM-GS, supporting the benefit of explicit conflict detection.

\section{Conclusion}
This work investigated the role of gradient conflicts in PINNs and their impact on optimisation dynamics. 
Through an empirical analysis of multiple PINNs benchmarks, we showed that angle-based and magnitude-based conflicts exhibit an alternating pattern throughout optimisation, with the manifestation of one conflict type coinciding with a reduced occurrence of the other.
These observations suggest that treating both phenomena similarly, as done by many existing gradient surgery methods, is suboptimal for PINNs.
Motivated by these findings, we proposed PAM-GS, a gradient surgery method that explicitly accounts for the different nature of gradient conflicts encountered during PINN optimisation. 
Overall, our results highlight the importance of characterising gradient conflicts throughout training rather than treating them as a static phenomenon.
By selecting its behaviour according to the detection of specify gradient conflict types, PAM-GS leads to a more effective and balanced optimisation over all tasks.

\bibliographystyle{unsrt} 
\bibliography{references}  

\appendix
\section{Partial Differential Equation Benchmarks} \label{appendix:PDE specification}
\paragraph{1D Burgers Equation.}
The Burgers equation is a nonlinear model frequently used to study the formation and propagation of shock waves. 
For a scalar field $u(x,t)$, the governing equation in one spatial dimension is
\begin{equation}
\frac{\partial u}{\partial t} + u \frac{\partial u}{\partial x} = \nu \frac{\partial^2 u}{\partial x^2},
\label{eq:burgers}
\end{equation}
where the viscosity coefficient is fixed at $\nu = 0.01/\pi$. 
The computational domain is defined by $t \in [0,1]$ and $x \in [-1,1]$, together with
$u(x,0) = -\sin(\pi x)$ and $u(-1,t) = u(1,t) = 0$.

For this configuration, no closed-form analytical solution is available. 
Therefore, numerical solutions generated using \textit{PhiFlow} \cite{Holl2020_iclr} are adopted as reference data for assessing the performance of PINNs.

\paragraph{1D Schr\"odinger Equation.}
The nonlinear Schr\"odinger equation describes the evolution of a complex-valued field
$h(x,t)=u(x,t)+iv(x,t)$ and is given by
\begin{equation}
i\frac{\partial h}{\partial t} + \frac{1}{2}\frac{\partial^2 h}{\partial x^2} + |h|^2 h = 0,
\label{eq:schrodinger}
\end{equation}
on the domain $x \in [-5,5]$ and $t \in [0,\pi/2]$. Periodic boundary conditions are imposed, while the initial state is $h(x,0)=2\,\mathrm{sech}(x).$
Since an analytical solution is not available under these conditions, numerical results obtained from \textit{DeepXDE} \cite{Bajaj2023_mlst} are employed as the benchmark solution for evaluating the trained PINNs.

\paragraph{2D Kovasznay Flow and 3D Beltrami Flow.}
Both Kovasznay and Beltrami flows represent exact solutions of the incompressible Navier--Stokes equations,
\begin{align}
\frac{\partial \mathbf{u}}{\partial t} + (\mathbf{u}\cdot\nabla)\mathbf{u} &= -\nabla p + \frac{1}{Re}\nabla^2\mathbf{u}, \quad
\nabla\cdot\mathbf{u} =0,
\label{eq:navier_stokes}
\end{align}
where $\mathbf{u}$ denotes the velocity field, $p$ the pressure, and $Re$ the Reynolds number.

\textbf{Kovasznay Flow.}
Kovasznay flow corresponds to a steady-state two-dimensional solution, implying
$\partial \mathbf{u}/\partial t = 0$. 
The exact velocity field $\mathbf{u}=[u_x,u_y]$ and pressure $p$ are  
\[(u_x(x,y) = 1 - e^{\lambda x}\cos(2\pi y),\quad u_y(x,y) = \frac{\lambda}{2\pi} e^{\lambda x}\sin(2\pi y),\quad p(x,y) = \frac{1}{2}\left(1-e^{2\lambda x}\right),
\]
with $\lambda=\frac{1}{2\nu} - \sqrt{\frac{1}{4\nu^2}+4\pi^2}$, $\nu=\frac{1}{Re}=\frac{1}{40},$ and  $(x,y) \in [-0.5,1] \times [-0.5,1.5]$.

\textbf{Beltrami Flow.}
Beltrami flow is an unsteady three-dimensional solution of the Navier--Stokes equations. 
The velocity components $\mathbf{u}(x,y,z,t)=[u_x,u_y,u_z]$ and pressure field $p(x,y,z,t)$ are expressed as:
\begin{align*}
u_x &= -a \Big[ e^{ax}\sin(ay+dz) + e^{az}\cos(ax+dy) \Big] e^{-d^2 t}, \\
u_y &= -a \Big[ e^{ay}\sin(az+dx) + e^{ax}\cos(ay+dz) \Big] e^{-d^2 t},\\
u_z &= -a \Big[ e^{az}\sin(ax+dy) + e^{ay}\cos(az+dx) \Big] e^{-d^2 t},\\
p &= -\frac{a^2}{2} \Big[e^{2ax}+e^{2ay}+e^{2az}+2\,e^{a(y+z)} \sin(ax+dy)\cos(az+dx)\\
&+2\,e^{a(z+x)} \sin(ay+dz)\cos(ax+dy)+2\,e^{a(x+y)} \sin(az+dx)\cos(ay+dz) \Big] e^{-2d^2 t}.
\label{eq:beltrami_pressure}
\end{align*}
For the present study, $a=d=1$, corresponding to $Re=1$. 
The computational domain is \(x,y,z \in [-1,1],\) and \(t \in [0,1].\)
Dirichlet boundary conditions are prescribed for both Kovasznay and Beltrami flows using values obtained from the analytical solutions. 
In addition, the initial condition for the Beltrami flow is also specified directly from the exact solution.

\section{Training Details}\label{appendix:Training Details}

We follow the standard PINN training protocol of \cite{liu2025_iclr}. Each model is a fully connected network with four hidden layers of 50 neurons, $\tanh$ activations, and Xavier initialisation. Training points are generated via Latin hypercube sampling and resampled at each iteration.
For the extended-training setting, we use a fixed learning rate of $10^{-4}$; otherwise, we use cosine annealing from $10^{-3}$ to $10^{-4}$ with a 100-epoch warm-up. All methods are optimised with Adam ($\beta_1=0.9$, $\beta_2=0.999$, $\epsilon=10^{-8}$). The number of collocation points for interior, boundary, and initial-condition points $(n_N,n_B,n_I)$ are $(10{,}000,250,250)$ for 1D Burgers, $(20{,}000,500,500)$ for 1D Schr\"odinger, $(20{,}000,1000,-)$ for 2D Kovasznay, and $(25{,}000,5000,5000)$ for 3D Beltrami flow. 
The corresponding training budgets are $3\times10^4$ epochs for 1D Burgers and $10^5$ epochs for all other problems.
Hyperparameters are selected via Bayesian optimization. \cite{dewancker2016bayesian}. 
We find that the optimal hyperparameters are consistent across benchmarks for CAGrad ($c=0.4$), FAMO ($\gamma=10^{-4}$, $lr_w=0.025$), DWA ($T=2$), and SAM-GS ($\gamma=0.9$, $\beta_1=0.9$, $\beta_2=0.99$). 
PAM-GS similarly uses fixed $\beta_1=0.9$ and $\beta_2=0.99$, with only $\gamma$ varying across benchmarks: $0.6$, $0.4$, $0.4$, and $0.7$ for 1D Burgers, 1D Schr\"odinger, 2D Kovasznay, and 3D Beltrami, respectively.

\paragraph{Runtime and Memory Analysis.} We measure runtime and memory overhead for three independent training of 2D Kovasznay on an NVIDIA H200 GPU.
In Table \ref{tab:memory_time}, we report the mean memory usage (in MB) and training time (in seconds) over 1,000 training steps. Overall, no substantial differences in computational or memory requirements are observed across the considered methods. Although PAM-GS maintains additional momentum statistics, its memory and computational requirements remain comparable to those of the other methods.

\begin{table}
\centering
\label{tab:memory_time}
\footnotesize 
\setlength{\tabcolsep}{2.5pt} 
\renewcommand{\arraystretch}{0.85}
\begin{tabular}{lcc}
\toprule
Method & Memory (MB) & Time (s) \\
\midrule
DWA         & 884.25  & $18.375 \pm 0.187$ \\
FAMO        & 884.26  & $18.616 \pm 0.023$ \\
ADMA        & 884.25  & $18.971 \pm 0.961$ \\
GradDrop    & 1605.13 & $23.588 \pm 0.917$ \\
SAM-GS      & 1605.33 & $24.220 \pm 0.277$ \\
PAM-GS      & 1605.33 & $24.281 \pm 0.085$ \\
ConFIG      & 1604.90 & $24.347 \pm 0.379$ \\
UW-SO       & 1605.13 & $24.371 \pm 0.032$ \\
Aligned-MTL & 1605.17 & $24.581 \pm 0.630$ \\
PCGrad      & 1604.91 & $25.247 \pm 1.115$ \\
CAGrad      & 1605.13 & $25.749 \pm 0.911$ \\
IMTLg       & 884.25  & $28.142 \pm 0.173$ \\
Nash-MTL    & 884.25  & $28.202 \pm 0.588$ \\
\bottomrule
\end{tabular}
\caption{Table B: }{Peak GPU memory and training time of the evaluated methods.}
\end{table}

\section{Comprehensive Results for the 3D Beltrami Flow}\label{appendix:3DBeltrami}
\begin{table}[h]
\resizebox{\textwidth}{!}{ 
\begin{tabular}{llllllllllllllllllllllllll}

\toprule
&
\multicolumn{4}{c}{Pressure $p$} &&
\multicolumn{4}{c}{Velocity $u$} &&
\multicolumn{4}{c}{Velocity $v$} &&
\multicolumn{4}{c}{Velocity $w$} &&
&&& \\
\cmidrule(lr){2-6}
\cmidrule(lr){7-11}
\cmidrule(lr){12-16}
\cmidrule(lr){17-21}

Method &&
BC & IC & Interior & MSE &&
BC & IC & Interior & MSE &&
BC & IC & Interior & MSE &&
BC & IC & Interior & MSE &&
Total MSE &&
MR & $\Delta M$ (\%) \\
\midrule
Base Line &  & 2.214e+01 & 1.865e+01 & 1.565e+01 & 1.946e+01 &  & 4.275 & 2.262 & 2.116 & 3.211 &  & 4.802 & 2.915 & 2.014 & 3.571 &  & 4.085 & 2.292 & 1.376 & 2.897 &  & 7.284 &  & NA & 0.000 \\
\midrule
UW-SO &  & 318400 & 2034000 & 124700 & 602200 &  & 79230 & 211600 & 46930 & 95830 &  & 90850 & 230700 & 52900 & 107200 &  & 83540 & 218200 & 45930 & 98950 &  & 226000 &  & 10.90 & $>1000$ \\
FAMO &  & 341000 & 1207000 & 337600 & 513200 &  & 76920 & 181600 & 36550 & 85500 &  & 80500 & 187200 & 34910 & 87890 &  & 79330 & 185300 & 35110 & 86990 &  & 193400 &  & 10.10 & $>1000$ \\
DWA &     & 35.96 & 29.28 & 28.96 & 32.48 &  & 8.753 & 5.761 & 2.753 & 6.318 &  & 9.972 & 7.138 & 2.806 & 7.212 &  & 8.827 & 5.244 & 2.982 & 6.321 &  & 13.08 && 9.75 & 92.50\\ 
IMTLg &  & 34.22 & 18.64 & 41.91 & 33.46 &  & 5.462 & 1.806 & 2.496 & 3.823 &  & 4.524 & 1.833 & 3.015 & 3.524 &  & 6.112 & 2.202 & 2.480 & 4.218 &  & 11.26 &  & 8.1 & 36.71 \\
Nash-MTL  &  & 23.08 & 17.01 & 19.17 & 20.67 &  & 3.910 & 2.170 & 1.883 & 2.942 &  & 4.336 & 2.802 & 1.523 & 3.168 &  & 4.299 & 2.374 & 1.205 & 2.967 &  & 7.436 &  & 6.2 & -2.491 \\
GradDrop &  & 21.47 & 16.19 & 14.59 & 18.31 &  & 5.421 & 2.236 & 2.163 & 3.787 &  & 5.620 & 2.009 & 3.005 & 4.097 &  & 4.302 & 2.367 & 1.554 & 3.074 &  & 7.316 &  & 5.75 & 4.811 \\
PCGrad &  & 17.93 & 15.39 & 14.75 & 16.45 &  & 4.908 & 2.517 & 2.078 & 3.563 &  & 5.497 & 2.617 & 2.681 & 4.059 &  & 5.766 & 2.517 & 2.381 & 4.080 &  & 7.038 &  & 6.5 & 9. \\
CONFIG & & 21.51 & 16.82 & 14.94 &  18.56 & & 3.077 & 1.314 & 2.268 &   2.477 & &4.844 & 1.569 & 4.069 &  3.951 & & 2.843 & 1.571 & 1.602 & 2.209 & & 6.799&  & 5.2 & -6.791\\
CAGrad &  & 20.95 & 13.38 & 19.66 & 19.04 &  & 3.180 & 1.541 & 2.061 & 2.510 &  & 3.182 & 1.381 & 1.363 & 2.265 &  & 3.886 & 1.652 & 1.222 & 2.624 &  & 6.609 &  & 4.0 & -16.89 \\
Aligned-MTL  &  & \textbf{12.77} & \textbf{8.946} & 11.96 & \textbf{11.76} &  & 3.827 & 1.377 & 2.387 & 2.896 &  & 4.675 & 1.510 & 2.448 & 3.360 &  & 4.424 & 1.859 & 2.783 & 3.409 &  & 5.355 &  & 3.15 & -11.79 \\
SAM-GS &  & 20.67 & 13.11 & 12.46 & 16.64 &  &  2.742 &   1.657 &  1.078 &   2.016 &  & 3.064 &   1.426 &  1.954 &   2.397 &  &  3.864 &   1.513 &  2.512 &   2.980 &  &  6.009 &  &  3.5 &  -18.36 \\
\midrule
PAM-GS &  & 18.18 & 8.987 & \textbf{10.97} & 14.14 &  & \textbf{2.351} & \textbf{0.823} & \textbf{0.900} & \textbf{1.602} &  & \textbf{2.536} & \textbf{0.798} & \textbf{1.144} & \textbf{1.762 }&  & \textbf{1.794} & \textbf{0.818} & \textbf{0.728} & \textbf{1.273} &  & \textbf{4.693} &  & \textbf{1.200} & \textbf{-46.13} \\
\bottomrule
\end{tabular}
}
\caption{Performance comparison of GS methods on the 3D Beltrami flow. All MSE values are reported in units of $10^{-5}$. Lower values indicate better performance.}
\label{table:3DBeltramiFull}
\end{table}

\end{document}